\documentclass[conference]{IEEEtran}
\usepackage{times}

\usepackage[numbers]{natbib}
\usepackage{multicol}
\usepackage[bookmarks=true]{hyperref}
\usepackage{caption}
\usepackage{amsmath,amssymb,amsfonts}
\usepackage{algorithm}
\usepackage{algorithmic}
\usepackage{graphicx}
\usepackage{textcomp}
\usepackage{booktabs}
\usepackage{multirow}
\usepackage{colortbl} 
\usepackage{xcolor}   
\usepackage{pifont}

\definecolor{lightblue}{RGB}{230, 242, 255}
\definecolor{lightgray}{RGB}{240,240,240}    
\definecolor{red}{RGB}{255,0,0}
\definecolor{green}{RGB}{0,128,0}

\usepackage[table]{xcolor} 
\usepackage{gradient-text}
\usepackage{xspace}
\usepackage{marvosym}
\newcommand\NickName{{\gradientRGB{VQ-Touch}{255, 106, 0}{192, 0, 0}}\xspace}

\def\BibTeX{{\rm B\kern-.05em{\sc i\kern-.025em b}\kern-.08em
    T\kern-.1667em\lower.7ex\hbox{E}\kern-.125emX}}

\author{
Kailin Lyu$^{1,2}$ \quad
Long Xiao$^{1}$ \quad
Jianing Zeng$^{1}$ \quad
Di Wu$^{1}$ \quad
Lin Shu$^{1}$ \quad
Jie Hao$^{1}$\\[4pt]
$^{1}$Institute of Automation, Chinese Academy of Sciences\\
$^{2}$Zhongguancun Academy
}

\begin{document}

\title{\raisebox{-1.8mm}{\includegraphics[width=1.5cm]{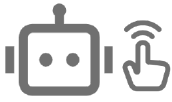}}~\textbf{\textit{\NickName}}: A Data-Efficient Tactile Generation Framework Across Sensors and Scenarios}

\maketitle

\begin{abstract}
Tactile image generation significantly reduces the dependency on expensive and wear-prone sensors by synthesizing high-fidelity tactile data, offering an efficient solution for tactile information acquisition in robotic perception and human-machine interaction systems. However, existing methods depend on large-scale, diverse datasets from specific sensors and lack efficient data utilization and robust generalization capabilities, struggling in vision-limited environments. To address this, we introduce \textbf{\textit{\NickName}}, a tactile generation framework that supports both cross-sensor and multi-scenario applications. Specifically, to efficiently extract complex deformation and texture features from the data, we propose DM-VQGAN, an effective tactile representation learner. Furthermore, we introduce a discrete diffusion decoder with a unified conditioning interface, supporting multimodal generation tasks such as images and labels, and enhances the model’s generalization capability through few-shot mixed training, thus achieving compatibility with current mainstream sensors and their variants. Experiments show that \textbf{\textit{\NickName}} surpasses state-of-the-art methods in multiple tasks.
\end{abstract}

\IEEEpeerreviewmaketitle

\section{Introduction}
\label{sec:intro}

Touch is a critical modality for human interaction with the environment, providing rich information about geometry, texture, and contact forces, and is therefore increasingly exploited in robotic manipulation \cite{calandra2018more,sunil2023visuotactile}. Vision-based tactile sensors (VBTS) \cite{liu2022sensor}, which use a camera to capture fine-grained elastomer deformations at high resolution and low cost, have emerged as the dominant tactile sensing paradigm. However, their limited durability and the laborious real-world acquisition pipeline render tactile data substantially scarcer and noisier than visual data \cite{zhang2022hardware,lyu2026touchthinker}. This bottleneck has spurred growing interest in synthesizing high-fidelity tactile images as a scalable surrogate for physical collection, lowering the training cost of downstream tasks such as robotic manipulation \cite{cao2022manipulation1} and texture recognition \cite{acharya2024exploringr58,lyu2026touchformer,xiao2026tacexpert,lyu2026tacreasoner}.

\begin{figure}[!t]
    \centering
    \includegraphics[width=\linewidth]{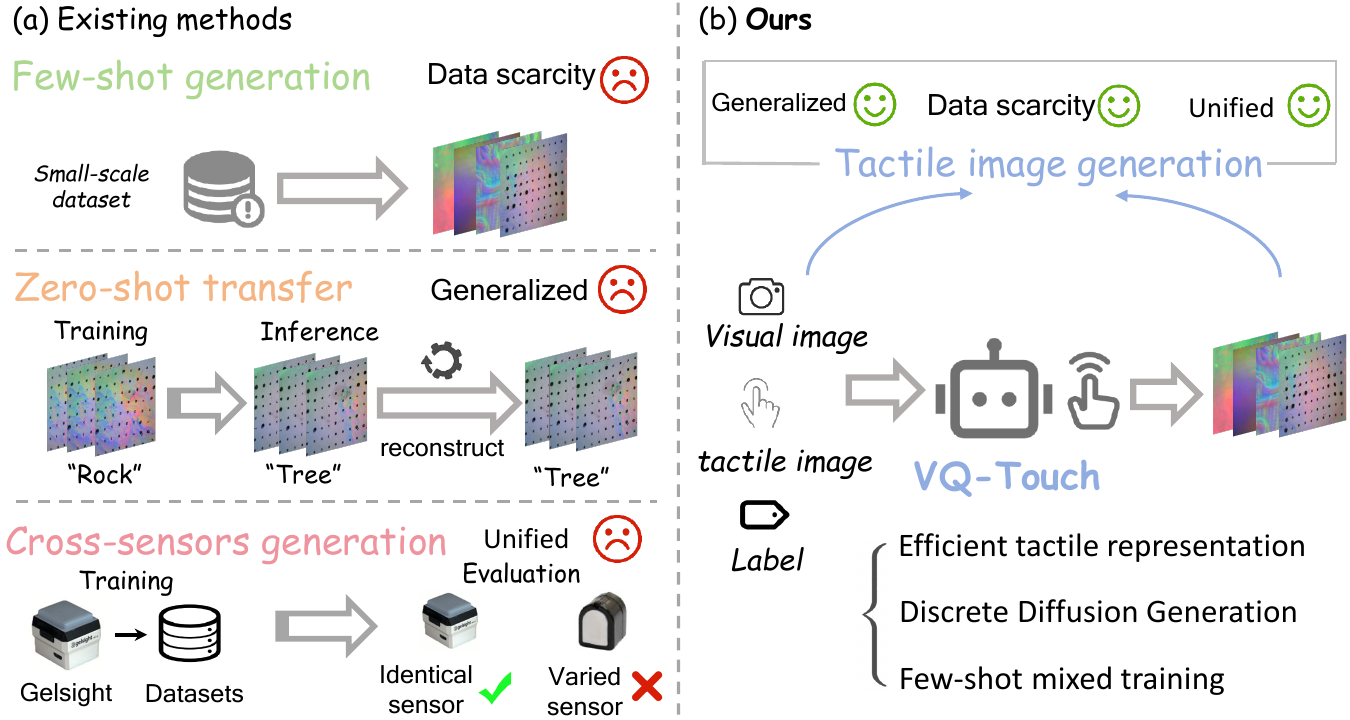}
    \caption{\textbf{The illustration of tactile generation tasks and comparison of different methods.} Leveraging data-efficient tactile representations and a unified discrete diffusion model, \textbf{\textit{\NickName}} supports few-shot, zero-shot transfer, and cross-sensor tactile generation.}
    \label{fig:intro fig}
    \vspace{-3mm}
\end{figure}

The fundamental paradigm of tactile image generation is to first extract diverse tactile feature representations from existing data and then synthesize tactile images that satisfy specific scene requirements. Although previous work has made significant progress \cite{lygerakis2024manipulation4}, two challenges remain: \ding{182} the need for large-scale and diverse datasets for pretraining each sensor type \cite{yang2022touch,zhao2024transferabler56}, which results in significantly degraded performance in environments where tactile data acquisition is difficult; and \ding{183} the inability of learned features to generalize to image generation tasks involving other sensors. This naturally leads to the following question: \textbf{\textit{Can we learn a data-efficient tactile representation that facilitates tactile generation across sensors and scenarios?}}

To address this challenge, we introduce \textbf{\textbf{\textit{\NickName}}}, a tactile generation framework that supports cross-sensor and multi-scenario applications. First, to model the low-variance latent space of tactile images while preserving their rich informational content, we adopt a VQGAN-based architecture, incorporating deformable convolutions and a multi-scale fusion module to extract both macro-level deformation patterns and micro-level texture features from tactile images. This enhances its applicability to vision-based tactile sensors (VBTS). Additionally, we propose a few-shot mixed training approach that enables the transfer of learned tactile representations from existing sensor data to image generation tasks for new sensors, facilitating data-efficient cross-sensor generalization. Based on these learned representations, we further train a discrete diffusion model to support multiple input modalities and perform conditional generation of corresponding tactile images. Our contributions are summarized as follows:

\begin{itemize}
    \item We introduce \textbf{DM-VQGAN}, an efficient tactile representation learner that captures generalizable, information-rich features tailored to tactile data.
    \item We propose a \textbf{unified multimodal conditional generation framework} that supports multiple input modalities, including tactile images, visual images, and semantic labels, to address diverse tactile generation requirements.
    \item We model tactile sensors at the sensor-family level and use clustering and \textbf{few-shot mixed training} to transfer features from one sensor to its family, eliminating full multi-dataset training.
    \item \textbf{\textit{\NickName}} outperforms state-of-the-art models in tactile image reconstruction and generation, excelling in vision-limited scenarios.
\end{itemize}

\begin{figure*}[!htbp]
    \centering
    \includegraphics[width=0.9\linewidth]{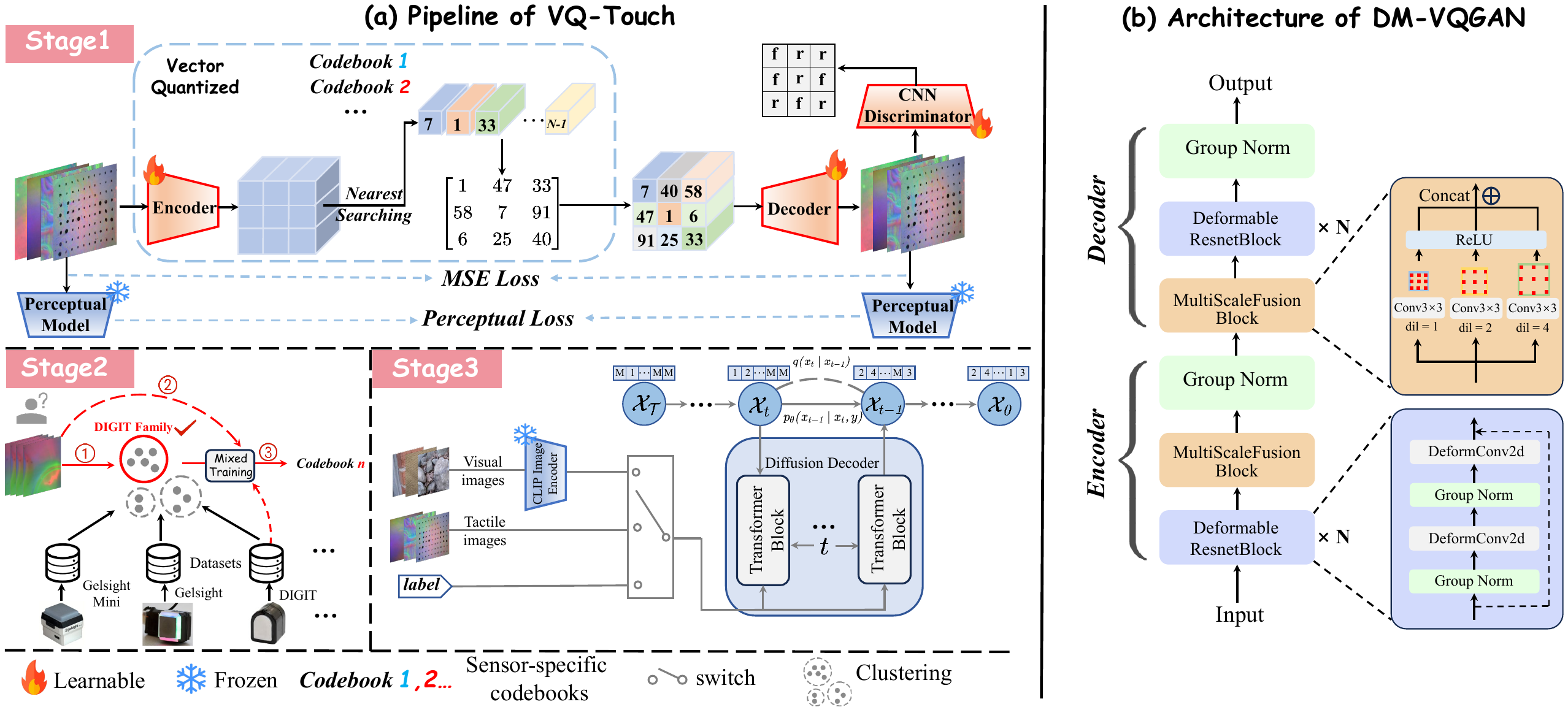}
    \caption{\textbf{Overview of the proposed \textbf{\textit{\NickName}} framework.} The framework consists of three stages: \textbf{(1)} learning efficient tactile features using the DM-VQGAN encoder-decoder architecture, which is based on discrete codebooks for representation, \textbf{(2)} acquiring sensor-family-specific codebooks through clustering and few-shot mixed training method, and \textbf{(3)} by leveraging the discrete latent space and encoder learned by DM-VQGAN, a discrete diffusion decoder is employed to generate high-quality tactile images conditioned on unimodal inputs including visual images, tactile images or semantic labels.}
    \label{fig:pipline}
    \vspace{-3mm}
\end{figure*}

\section{Methodology}
\label{method}
\noindent \textbf{Overview.} In this section, we elaborate on \textbf{\textbf{\textit{\NickName}}}, whose pipeline is shown in Fig.\ref{fig:pipline}. It consists of three components: \textbf{(i)} \textbf{DM-VQGAN}, a tactile-specific encoder that models and amplifies geometric variations to produce generation-friendly latent representations (\textbf{Sec.\ref{3.2}}); \textbf{(ii)} a \textbf{Few-shot Mixed Training} scheme that transfers these representations across sensor families for data-efficient cross-sensor generalization (\textbf{Sec.\ref{3.2}}); and \textbf{(iii)} a unified \textbf{discrete-diffusion} generative framework that conditions on multimodal inputs and uses the pretrained encoder to synthesize tactile images (\textbf{Sec.\ref{3.4}}).


\subsection{DM-VQGAN}
\label{3.2}



Vision-based tactile images discard color while preserving essential cues such as morphology, geometry, and force distribution \cite{li2013sensingr68}. Exploiting this low intrinsic variance together with VQGAN's strong latent-space modeling \cite{cao2023efficient}, we adopt a discrete codebook to obtain structured representations under limited data. Vanilla VQGAN, however, relies on fixed convolutions and thus struggles to capture the non-rigid deformations and multi-scale patterns characteristic of VBTS \cite{li2013sensingr68,dai2017deformable}. We therefore propose \textbf{DM-VQGAN}, which integrates \underline{d}eformable and \underline{m}ulti-scale dilated convolutions to strengthen feature extraction and representation, as shown in Fig.~\ref{fig:pipline}(b).

\textbf{Training Process of DM-VQGAN.} Similar to the classic VQGAN \cite{esser2021taming}, DM-VQGAN consists of an encoder \(E\), a decoder \(G\), a patch-based discriminator \(D\), and a codebook \(Z = \{ z_k \}_{k=1}^{K}\), containing \(K\) discrete codes, as shown in Stage1 of Fig.~\ref{fig:pipline}(a). Given an input image \(x \in \mathbb{R}^{H \times W \times 3}\), the encoder \(E\) first extracts a latent feature representation \(\hat{z} \in \mathbb{R}^{H_f \times W_f \times n_z}\), where \(n_z\) and \(f\) represent the dimensionality of the latent features and the spatial compression ratio, respectively. Next, the feature vector at each spatial location \((i, j)\) is quantized to the closest discrete code from the codebook via nearest neighbor search:
\begin{equation}
z_q = q(\hat{z}) := \left( \arg \min_{z_k \in Z} \| \hat{z}_{ij} - z_k \| \right) \in \mathbb{R}^{H_f \times W_f \times n_z},
\label{eq6}
\end{equation}
The decoder \(G\) then decodes the quantized features back into the image space, i.e., \(\hat{x} = G(z_q)\). The training objective of DM-VQGAN is to minimize the difference between the input image and the reconstructed image, and the corresponding loss function can be expressed as:
\begin{equation}
\begin{aligned}
L_{\text{total}} =\; & 
\underbrace{
\| x - \hat{x} \|_1 
+ \| \text{sg}[E(x)] - z_q \|_2^2 
+ \beta \| \text{sg}[z_q] - E(x) \|_2^2
}_{L_{\text{vq}}} \\
& + L_{\text{per}} + L_{\text{adv}},
\end{aligned}
\label{eq7}
\end{equation}
where \(L_{per}\) and \(L_{adv}\) represent the perceptual loss and the adversarial loss. In the reconstruction loss \(L_{vq}\), \(\text{sg}[\mathpunct{\cdot}]\) denotes the stop-gradient operation, and \(\beta \| \text{sg}[z_q] - E(x) \|_2^2\) is known as the commitment loss, where the commitment weight \(\beta\) is set to 0.25 \cite{goodfellow2020generative, yu2021vector}.

\textbf{Deformation Pattern Extraction}: In each encoder and decoder, we introduce deformable convolutions \cite{dai2017deformable} to adaptively learn dynamic spatial offsets for convolution kernels, enabling the model to better capture local, complex, and irregular deformation patterns present in tactile images. The structure of the deformable convolution layer, denoted as:
\begin{equation}
y(p_0) = \sum_{p_k \in \mathcal{G}} w_k \cdot x \left( p_0 + p_k + \Delta p_k \right),
\label{eq8}
\end{equation}
where \(y(p_0)\) is the output feature vector at position \(p_0\), and \(p_k \in \mathbb{R}^2\), \(\Delta p_k \in \mathbb{R}^2\) represent the pre-defined offset and learned offset for the \(k\)-th sampling point. This flexible convolutional sampling mechanism facilitates effective learning of dynamic spatial relationships and feature variations, thereby enhancing sensitivity to localized deformations during tactile representation learning. 

\textbf{Multi-scale Fusion}: To effectively capture the rich, multi-scale features inherent in tactile images, we design a multi-scale fusion module that employs parallel dilated convolutions with varying dilation rates (i.e., 1, 2, and 4), denoted as:
\begin{equation}
y = \mathcal{F}_{\text{conv3x3}} \left( \text{Concat} \left( \left\{ \sigma \left( \mathcal{F}_{\text{dilate}}^{(d_i)} (x) \right) \right\}_{d_i \in \{1, 2, 4\}} \right) \right).
\label{eq9}
\end{equation}

This design expands the receptive field, enabling the simultaneous extraction of macro-level global deformations and micro-level textural features. Consequently, the DM-VQGAN effectively addresses the limitations of traditional convolutional structures in representing complex dynamic deformations and multi-scale features of tactile images, enhancing tactile representation and improving tactile image generation.

\subsection{Few-shot Mixed Training for Cross-sensor Generalization}
\label{3.4}

Many studies \cite{yang2024binding,sun2025zongshu} have observed substantial discrepancies among tactile sensors, which hinder the generalization of tactile representation models. Existing methods, such as UniTouch \cite{yang2024binding} and T3 \cite{zhao2024transferabler56}, typically learn sensor-specific features from multiple large-scale datasets. Although effective, such a strategy is data-intensive and potentially redundant.

By analyzing mainstream tactile datasets \cite{yang2022touch,fu2024touch}, we identify two major sources of variation in tactile images: intrinsic sensor properties and environmental factors. The former are determined by sensor design and manufacturing and remain relatively stable, while the latter, including illumination, calibration, and material degradation, introduce appearance shifts that may obscure consistent tactile patterns. We therefore hypothesize that sensors \textit{$A_1$} and \textit{$A_2$} with shared intrinsic properties but different environmental conditions belong to the same sensor family $\textit{A}$, whose representations can be jointly learned. Consequently, a model trained on \textit{$A_1$} can be adapted to \textit{$A_2$} with only minimal additional data.

To identify the family of an unseen sensor, we propose an unsupervised clustering strategy. Specifically, DM-VQGAN is used to extract tactile features from images of multiple sensor types, followed by cross-sensor clustering. For each cluster, we compute its prototype vector by iteratively estimating feature centroids, and define the inter-cluster distance as
\begin{equation}
D(C_i, C_j) =
\left\|
\frac{1}{|C_i|} \sum_{x \in C_i} x
-
\frac{1}{|C_j|} \sum_{y \in C_j} y
\right\|, \quad i,j \in \mathbb{N},
\label{eq9}
\end{equation}
where \( |C_i| \) and \( |C_j| \) denote the numbers of samples in clusters \( C_i \) and \( C_j \), respectively, and \(x\) and \(y\) are their corresponding feature vectors. Given an unseen sensor, we compute its prototype vector and assign it to the nearest sensor family. We then construct a small-sample mixed dataset to train DM-VQGAN for tactile representation learning, avoiding full retraining on all datasets.

\subsection{Multimodal Conditional Generation Framework}
\label{3.3}

Although DM-VQGAN learns effective tactile representations, its original decoder cannot flexibly incorporate external prompts such as images. We therefore adopt discrete diffusion \cite{gu2022vqdiffusion}, which naturally operates in discrete latent spaces, to build a unified multimodal conditional generation framework. As shown in Stage 3 of Fig.~\ref{fig:pipline}(a), the diffusion decoder is conditioned on a single modality at each training step:
\begin{equation}
\boldsymbol{c} = f_{m}(y_{m}), \quad
m \in \{\text{\textit{visual}},\,\text{\textit{tactile}},\,\text{\textit{label}}\},
\label{eq10}
\end{equation}
where \(f_m(\cdot)\) denotes the encoder for modality \(m\), and \(\boldsymbol{c}\) is the corresponding conditional latent feature. This design enables flexible tactile image generation from visual, tactile, or label conditions according to task requirements. For visual conditioning, we use a frozen pretrained CLIP encoder \cite{radford2021learning}, i.e.,
\(\boldsymbol{c}_{\text{visual}} = f_V^{\text{CLIP}}(y_{\text{visual}})\). For tactile conditioning, the input \(y_{\text{tactile}}\) is mapped to discrete indices using the sensor-specific codebook \(\mathcal{Z}=\{z_k\}_{k=1}^{K}\) learned by DM-VQGAN:
\begin{equation}
\boldsymbol{c}_{\text{tactile}} =
q\left(f_{\text{DM-VQGAN}}(y_{\text{tactile}}),\,\mathcal{Z}\right).
\label{eq12}
\end{equation}
The selected conditional feature \(\boldsymbol{c}\) then interacts with the noisy latent representation \(\boldsymbol{z}_t\) through transformer modules in the diffusion decoder, enabling condition-specific tactile synthesis.

\section{Experiments}
\label{experiments}

\subsection{Dataset and Baselines}

We train and evaluate the proposed model on four public tactile datasets, namely Touch and Go \cite{yang2022touch}, FabricVST \cite{cao2024multimodal}, VisGel \cite{var3}, and HCT \cite{fu2024touch}, covering GelSight, GelSight Mini, and DIGIT sensors. We compare our method with representative tactile generation approaches \cite{yang2024binding} across multiple downstream tasks. For fairness, all experiments are repeated three times, and the average results are reported.

\begin{table}[!t]
\centering
\caption{\textbf{Tactile image reconstruction.} We compare tactile image reconstruction across different methods and sample sizes, evaluated by FID ($\downarrow$) and SSIM ($\uparrow$).}
\label{table2}
\resizebox{\columnwidth}{!}{

\begin{tabular}{c c c c c c c c}
\toprule
\textbf{Method} & \multicolumn{3}{c}{\textbf{FID($\downarrow$)}} & \multicolumn{3}{c}{\textbf{SSIM($\uparrow$)}} & \multirow{2}{*}{\textbf{Mask Ratio}} \\ \cmidrule(lr){2-4}\cmidrule(lr){5-7}
\textbf{Sample Size} & \textbf{100} & \textbf{1K} & \cellcolor{lightgray}\textbf{10K} & \textbf{100} & \textbf{1K} &  \cellcolor{lightgray}\textbf{10K} &  \\ \midrule
\multirow{3}{*}{\textbf{MAE}} 
& 160.11 & 129.45 & \cellcolor{lightgray}117.42 & 0.680 & 0.695 & \cellcolor{lightgray}0.712 & \textbf{0.25} \\
& 205.32 & 209.22 & \cellcolor{lightgray}186.11 & 0.561 & 0.635 & \cellcolor{lightgray}0.645 & \textbf{0.5} \\
& 270.04 & 235.10 & \cellcolor{lightgray}220.86 & 0.480 & 0.582 & \cellcolor{lightgray}0.598 & \textbf{0.75} \\ 
\hline\hline
\addlinespace[2pt]
\textbf{VQGAN} & 46.76 & 18.34 & \cellcolor{lightgray}16.54 & 0.823 & 0.911 & \cellcolor{lightgray}0.923 & \textbackslash{} \\ 
\hline\hline
\addlinespace[2pt]
\textbf{DM-VQGAN} & 46.02 & 17.25 & \cellcolor{lightblue}\textbf{15.62} & 0.831 & 0.917 & \cellcolor{lightblue}\textbf{0.936} & \textbackslash{} \\ 
\bottomrule
\end{tabular}
}
\end{table}

\subsection{Results}
\label{4.2}
\begin{figure}[!t]
    \centering
    \includegraphics[width=\linewidth]{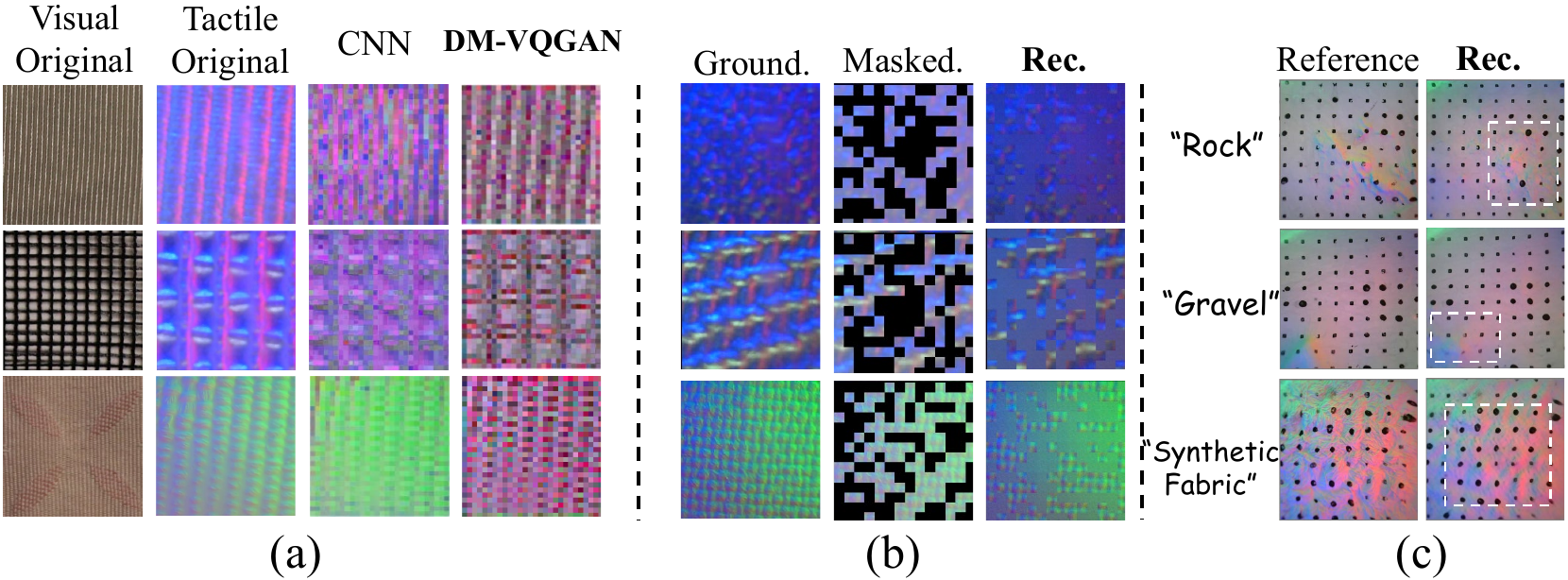}
    \caption{\textbf{Latent space visualization and reconstruction comparison.} \textbf{(a)} shows RGB latent features from three encoders. \textbf{(b)} presents the reconstruction results of MAE with a ViT-Base backbone. \textbf{(c)} shows zero-shot reconstruction by \textbf{\textit{\NickName}}, demonstrating generalization to unseen objects.}
    \label{fig:Figure/cnnvqgan_latent_mae}
\end{figure}

\noindent \textbf{Effectiveness Analysis of DM-VQGAN for Tactile Representation Learning.} To evaluate the effectiveness of DM-VQGAN for tactile representation learning, we conduct reconstruction experiments on the FabricVST dataset~\cite{cao2024multimodal} under varying training-sample scales, using MAE~\cite{he2022masked} and VQGAN~\cite{esser2021taming} as baselines. As summarized in Table~\ref{table2}, DM-VQGAN consistently achieves superior reconstruction quality, preserving fine tactile structures across all data scales. In contrast, MAE fails to recover detailed textures (Fig.~\ref{fig:Figure/cnnvqgan_latent_mae}(b)), and its performance with $N=10\mathrm{k}$ remains inferior to DM-VQGAN trained with only $N=100$ samples. Furthermore, latent-space visualization (Fig.~\ref{fig:Figure/cnnvqgan_latent_mae}(a)) shows that, unlike CNN autoencoders that mainly retain color distributions, DM-VQGAN learns more structured and compact representations that emphasize salient tactile features while suppressing background variations. This demonstrates its effectiveness in learning discriminative and low-variance tactile representations.

\begin{table}[!t]
\caption{\textbf{Quantitative Results on Tactile Generation Task.} Our method achieves the best results on most metrics.} 
\centering
\resizebox{\linewidth}{!}{
    \begin{tabular}{ccccccccc} 
    \toprule 
    \multirow{2}{*}{Method} & \multicolumn{4}{c}{HCT} & \multicolumn{4}{c}{SSVTP} \\
    \cmidrule(lr){2-5} \cmidrule(lr){6-9}
    & CTTP $\uparrow$ & LPIPS $\downarrow$ & SSIM $\uparrow$ & PSNR $\uparrow$ 
    & CTTP $\uparrow$ & LPIPS $\downarrow$ & SSIM $\uparrow$ & PSNR $\uparrow$ \\
    \midrule 
    GVST   & -     & 0.573 & 0.881 & 19.45 & -     & 0.502 & 0.918 & 21.15 \\
    UniTouch   & 0.156 & 0.528 & 0.902 & 19.84 & 0.127 & 0.555 & 0.824 & 12.42 \\
    PixArt-$\alpha$ & 0.198 & 0.504 & 0.876 & 20.26 & 0.125 & 0.497 & 0.916 & 22.35 \\
    TextToucher  & 0.261 & 0.427 & 0.904 & 22.70 & 0.152 & 0.465 & 0.930 & 22.43 \\
    \cellcolor{lightblue}\textbf{Ours}   
                    & \cellcolor{lightblue}\textbf{0.278}     
                    & \cellcolor{lightblue}\textbf{0.216}     
                    & \cellcolor{lightblue}\textbf{0.910}    
                    & \cellcolor{lightblue}\textbf{24.35}     
                    & \cellcolor{lightblue}\textbf{0.158}     
                    & \cellcolor{lightblue}\textbf{0.231} 
                    & \cellcolor{lightblue}\textbf{0.932} 
                    & \cellcolor{lightblue}\textbf{25.56} \\
    \bottomrule 
    \end{tabular}
}
\vspace{-5mm}
\label{tab3}
\end{table}

\begin{figure}[!t]
    \centering
    \includegraphics[width=0.95\linewidth]{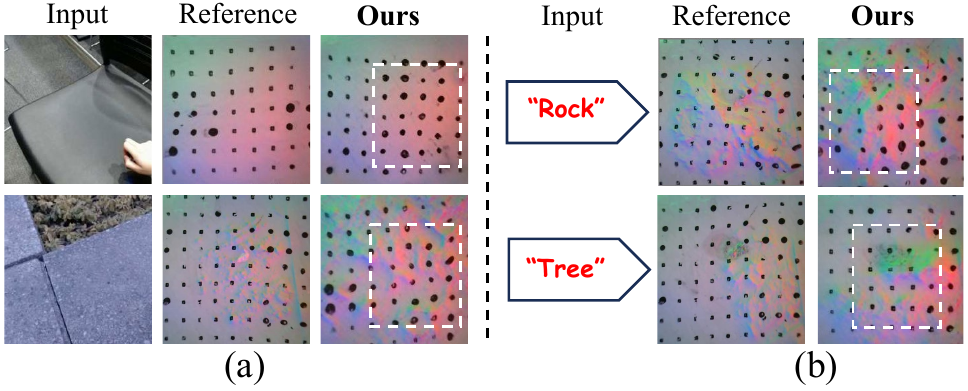}
    \caption{\textbf{Tactile image generation under vision or label conditions.} \textbf{\textit{\NickName}} can also generate high-fidelity tactile images from \textbf{(a)} vision inputs or \textbf{(b)} label inputs.}
    \label{fig:vision2touch}
\end{figure}

\noindent \textbf{Evaluation Across Multiple Tasks and Applications.} We evaluate the proposed framework across data-limited learning, multimodal generation, and downstream recognition tasks. Under limited samples and categories, our method demonstrates superior few-shot representation learning and zero-shot generalization to unseen objects, consistently outperforming competing approaches (Tab.~\ref{tab4}, Fig.~\ref{fig:Figure/cnnvqgan_latent_mae}(c)). We further assess vision- and label-conditioned tactile image generation using real-world scene images and semantic labels. As shown in Fig.~\ref{fig:vision2touch} and Tab.~\ref{tab3}, \textbf{\textit{\NickName}} produces tactile images that maintain strong cross-modal consistency, structural fidelity, and semantic alignment. Finally, classifiers trained on generated tactile images achieve higher object recognition accuracy than baseline methods (Tab.~\ref{down}), demonstrating the practical value of the generated data for downstream applications.

\begin{table}[t]
\centering
\caption{Performance of Tactile Images Generated by VQ-Touch on Object Recognition Tasks Across Different Datasets.}
\label{down}
\resizebox{0.48\textwidth}{!}{
\large
\begin{tabular}{*{6}{c}}
\toprule
Method & Dataset & Top-1 Acc (\%) & Precision (\%) & Recall & F1 score \\
\midrule
PixArt-$\alpha$ & TAG  & 81.26 & 81.8 & 0.82 & 0.819 \\
Ours & TAG  & \textbf{84.53} & \textbf{85.1} & \textbf{0.84} & \textbf{0.845} \\
PixArt-$\alpha$ & FabricVST  & 88.79 & 89.4 & 0.90 & 0.897 \\
Ours & FabricVST & \textbf{92.48} & \textbf{92.7} & \textbf{0.92} & \textbf{0.923} \\
\bottomrule
\end{tabular}
}
\end{table}

\begin{table}[!t]
\centering
\caption{\textbf{Comparison of few-shot tactile image reconstruction.}}
\label{tab4}
\resizebox{\linewidth}{!}{
\begin{tabular}{ccccc}
\toprule
method                    & FID($\downarrow$) & LPIPS($\downarrow$) & SSIM($\uparrow$) & Sample Size \\
\midrule
\multirow{2}{*}{Unitouch \cite{yang2024binding}} & 72.43  & 0.583    & 0.615   & N=500       \\
                          & 56.71  & 0.541    & 0.663   & N=2k        \\
\hline\hline
\multirow{2}{*}{Ours}     & 20.71  & 0.421    & 0.863   & N=500       \\
                          & 17.03  & 0.408    & 0.894   & N=2K        \\
\bottomrule
\end{tabular}}
\end{table}

\begin{figure}[!t]
    \centering
    \includegraphics[width=\linewidth]{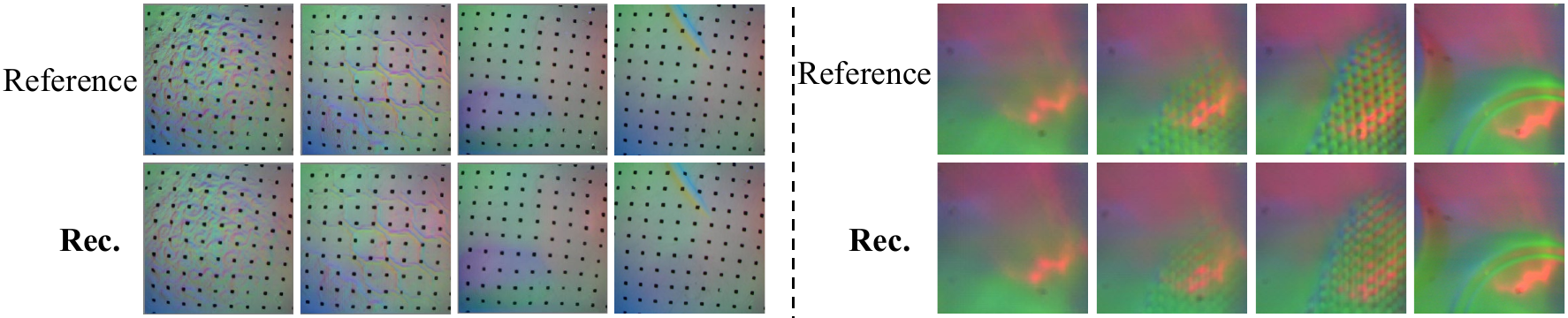}
    \caption{\textbf{Cross-Sensor Tactile Image Reconstruction. } We evaluate reconstruction performance by training with a few-shot mixed training method. The dataset consists of a small number of unseen sensor data (50 images from GelSight Derivative Version I and DIGIT) combined with data from their respective sensor families at a 1:5 ratio.}
    \label{fig:fulu_rec}
\end{figure}

\noindent \textbf{Validation of Transferability and Generalizability across Datasets and Sensors.} As discussed in Sec.~\ref{3.4}, we improve the generalizability of \textbf{\textit{\NickName}} across different sensors and datasets within the same sensor family through few-shot mixed training. Specifically, a small number of tactile images from unseen sensors are combined with existing datasets from their corresponding sensor families to train DM-VQGAN. As illustrated in Fig.~\ref{fig:fulu_rec}, the learned codebook can be directly transferred to tactile image reconstruction for unseen sensors, demonstrating the effectiveness of our approach in enabling cross-sensor generalization under limited-data settings.

\section{Conclusion}

In this work, we propose \textbf{\textit{\NickName}}, a novel and data-efficient framework for tactile image generation. \textbf{\textit{\NickName}} establishes a new paradigm for few-shot tactile generation by leveraging the representation learner DM-VQGAN and a unified conditional diffusion encoder, enabling cross-sensor compatibility and vision-free multimodal-to-tactile generation. Our framework demonstrates superior performance over SOTA methods while requiring limited training data. The proposed sensor-family clustering and mixed training strategy further enhance generalizability. These findings highlight the potential of \textbf{\textit{\NickName}} as an effective solution for scalable tactile generation and efficient tactile perception in robotics and human computer interaction applications.


\bibliographystyle{plainnat}
\bibliography{references}

\end{document}